# Tendon-Driven Soft Robotic Gripper for Berry Harvesting

Anthony L. Gunderman, Jeremy Collins, Andrea Myer, Renee Threlfall, Yue Chen

*Abstract*— **Global berry production and consumption have significantly increased in recent years, coinciding with increased consumer awareness of the health-promoting benefits of berries. Among them, fresh market blackberries and raspberries are primarily harvested by hand to maintain post-harvest quality. However, fresh market berry harvesting is an arduous, costly endeavor that accounts for up to 50% of the worker hours. Additionally, the inconsistent forces applied during hand-harvesting can result in an 85% loss of marketable berries due to red drupelet reversion (RDR). Herein, we present a novel, tendon-driven soft robotic gripper with active contact force feedback control, which leverages the passive compliance of the gripper for the gentle harvesting of blackberries. The versatile gripper was able to apply a desired force as low as 0.5 N with a mean error of 0.046 N, while also holding payloads that produce post-forces as high as 18 N. Field test results indicate that the gripper is capable of harvesting berries with minimal berry damage, while maintaining a harvesting reliability of 95% and a harvesting rate of approximately 4.8 seconds per berry.**

## I. INTRODUCTION

BERRY production and consumption have increased globally in recent years. According to a United States Department of Agriculture report, berries in the United States reached a market value of $7.5 billion in 2020 [1], and the global market is expected to grow an additional 9.3% by 2025 [2]. This rise in berry consumption is associated with the reported health-promoting benefits of fruits and fruit juices containing antioxidants [3-5], which can mitigate conversion of cellular macromolecules to specific reactive, oxidized forms [3, 4], a primary cause of chronic diseases [6, 7]. These fruits include blueberries, raspberries, and blackberries, which are grown by commercial and local growers.

However, as demand increases with market value, the greatest risk to the fresh market berry industry is a shortage of available skilled labor, which is strongly influenced by wage, local labor supply, and crop yield [8]. Harvesting for fresh markets is constrained to hand harvesting to prevent yield loss and damage. As a result, berry harvesting for fresh markets is an inherently labor-intensive operation, requiring a massive deployment of laborers, contributing up to 50% of the total

This research is funded by a University of Arkansas Chancellor's grant.

A. L. Gunderman, J Collins, and Y. Chen are with the Mechanical Engineering Department, University of Arkansas, Fayetteville, AR 72701 USA (corresponding Dr. Yue Chen; e-mail: algunder@uark.edu, yc039@uark.edu).

A. Myers and R. Threlfall are with the Food Science Department, University of Arkansas, Fayetteville, AR 72704 USA (corresponding Dr. Renee Threlfall; e-mail: rthrelf@uark.edu, alm028@uark.edu).

hours spent on the crop annually [8]. These costly high-intensity periods are short, with a 31-day annual harvesting window, and are the primary contributor to the average seasonal cost per blackberry tray (4.5 pounds) of $6.25 for the consumer [9]. However, failure to harvest meticulously can be detrimental to product quality, often resulting in berry disorders, such as red drupelet reversion (RDR) in blackberries [10]. RDR is a postharvest disorder where the drupelet changes color from black to red, diminishing marketability [11]. Current recommendations indicate that blackberry harvesting should be performed with minimal handling. However, without heeding such precautions, research demonstrates that up to 85% of the fruit can be afflicted with RDR [10]. Consequently, handling consistency are of paramount importance, but both are often impossible to control due to the immense variety in skill between harvesters. What is more, the COVID-19 pandemic has caused increased losses due to the hesitance of growers to permit large groups of day laborers on their farm, limiting the amount of fruit that can be harvested. As with most labor-intensive industries that require precision handling, there is an effort to integrate robotics into these tasks; however, existing harvesting machines are not suitable for harvesting delicate blackberries.

Current methods of automated harvesting rely on rough handling of the fruit by either (i) cutting the stem and catching the fruit in a separate vessel [12-17], (ii) shaking the fruit off of the plant [18, 19], (iii) picking the fruit with rigid components [20-22], or (iv) using compliant plastic grippers to pick the fruit [23-25]. For example, blueberries are typically harvested by shaking, and strawberries, oranges, apples, plums, and peppers are harvested using aforementioned methods (i), (iii), and (iv). However, each of these methods inevitably damage less robust berries, such as blackberries, which are a fragile, aggregate fruit comprised of drupelets surrounding a soft tissue receptacle or torus [26]. Furthermore, many methods using grippers are cumbersome, lack force feedback, and are limited to environmentally controlled greenhouse operations [23, 27, 28].

Soft robotics provides a novel option for automatic harvesting by leveraging the use of compliant grippers [29-32]. This is a departure from the aforementioned rigid robotic harvesting methods due to the inherent compliance of elastic materials used in the robotic body (rubber, silicone, etc.), enabling task versatility that is not found in traditional rigid-bodied robotic systems. These soft robotic systems are ideally suited for grasping [33-35] and manipulating delicate objects [36, 37] with complex, dynamic shapes. However, there are limitations associated with currently developed grippers with respect to their implementation in blackberry harvesting, such as: (i) they are too large, or not optimized, to be used inside the blackberry canopy [25, 33, 34, 36], (ii) they are too soft for use in the blackberry canopy [32, 37], or (iii) their semi-compliant plastic components still risk damage to the blackberry surface [23, 24]. Thus, the desire to create a robotic harvesting system (Fig. 1) that has an optimized gripper that can safely and efficiently harvest delicate blackberries, motivates the work in this project.

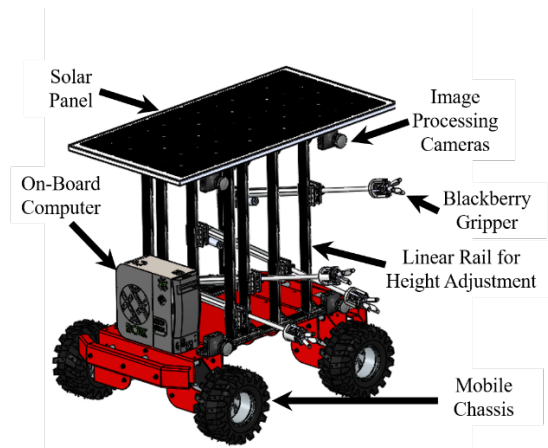

Fig 1. Proposed robotic system for soft gripper enabled berry harvesting. The soft gripper is implemented on a mobile chassis, which is able to identify, reach for, and grasp ripe berries. Note that multiple grippers can be used to obtain optimized harvesting efficiency vs. cost.

In this paper, we present a custom-designed, tendon-driven, soft robotic gripper for blackberry harvesting that utilizes force feedback to apply consistent handling forces. The tendon actuation method allowed a small cross-sectional footprint that provides increased dexterity within the blackberry canopy that cannot be obtained by other robotic harvesters. The gripper's tendon retraction was actively controlled using force feedback at the fingertips, which allowed harvesting consistency that surpasses the capabilities of the human hand. This robotic gripper was extensively validated through several metrics in a benchtop setting, as well as in a field test by harvesting 240 blackberries. A post-harvesting analysis was conducted on these berries to evaluate traits that impact berry marketability in order to compare the soft robotic gripper's harvesting ability with the standard hand harvesting method. Section II describes the soft gripper design, fabrication and prototyping, sensor integration, finger configuration, and the force feedback control method. Section III provides the experimental methods. Section IV elaborates on the results and a discussion of the preliminary field tests. The paper is concluded in Section V.

## II. Methods and Materials

### A. Design Requirements

Robotic grippers are generally designed with the goal of optimizing their performance to a specific task. The human hand, however, is well suited for a wide variety of tasks. In this paper, our gripper aims to outperform the human hand with respect to the specific task of blackberry harvesting. By limiting the scope of functions to strictly berry harvesting, the gripper design aims to overcome the limitations associated with hand harvesting of delicate berries by adhering to these constraints, which include:

1. The gripper must permit soft, accurate handling and manipulation of delicate caneberries. Current berry damage associated with hand harvesting indicates up to an 85% loss of marketable berries to red drupelet reversion [10]. In this preliminary study, it is desired to obtain similar, or less, berry damage percentages compared to that of hand harvesting methods.
2. The actuation method must be able to produce an adequate fingertip force of approximately 0.8 N to consistently pick berries from the cane using closed-loop force control.
3. The gripper must be able to reliably handle a large range of berry diameters and shapes (~1.5 to 5 cm), ensuring a harvesting reliability of at least 90%. Conversely, the gripper assembly must also have a small cross-sectional footprint to increase dexterity for navigation within the caneberry canopy.
4. The gripper must possess an appropriate amount of finger stiffness to prevent the fingers from being easily disturbed by the surrounding plant canes and leaves.
5. In order to implement this design as a compact, quiet, autonomous harvester in the future, the gripper assembly must remain untethered.

### B. Finger Design and Fabrication

Within soft robotics, there are several popular actuation methods: pneumatic [35-40], hydraulic [41], vacuum [42], tendon-driven [33, 43, 44], etc. [30, 31]. Although all listed methods meet design requirements 1 and 2, those that utilize fluid power, i.e. pneumatic, hydraulic, and vacuum, have several intrinsic properties that do not accommodate requirements 3-5. Fluid power, which requires a fluid tank and a compressor or evacuator, burdens the system with cumbersome components that require a tethered, high power, AC voltage source. What is more, pneumatics, vacuum, and hydraulics present increased cost and complexity in order to conform to IP ratings, as well as increased finger size due to fittings, integrated internal channels and transfer lines. Therefore, tendon-driven actuation was selected for this application, which requires only a motor for linear retraction and a 12V DC power supply.

Passive finger compliance was provided by casting the finger body out of a two-part silicone (Model: Dragon Skin FX Pro, Brand: Smooth-On), which permitted a soft interface for handling the berry. However, this property also leaves the fingers susceptible to external disturbance. To combat this drawback and meet design requirement 4, a 5 mm wide, 0.3 mm thick nitinol strip was used as an internal backbone, as shown in Fig. 2. This strip produced mildly increased stiffness in the direction of desired curvature, requiring a maximum tendon actuation force of only 20 N, but provided 278 × greater stiffness in the transverse direction.

Molding of the finger was accomplished using a two-part mold that was made of ABS plastic using additive manufacturing. This process began with the assembly of the internal structure. The nitinol backbone was adhered to a plastic tip and base using cyanoacrylate, as shown in Fig. 2. PTFE tubing was also attached to the plastic tip and base to create a cavity within the silicone mold for the tendon. The tendon, which was also included during the molding process, was a guitar string (36 gauge, Ernie Ball, CA USA) that was terminated in the upper plastic component with a 3 mm offset from the nitinol strip. This offset was used to provide eccentric loading during tendon retraction, resulting in inward bending of the finger. Once the finger backbone was assembled, the assembly was clamped between the lower and upper mold. The clamping force held the backbone assembly in place, while the nitinol strip provided the rigidity necessary to ensure the plastic tip was level with the plastic base. The silicone was poured through the opening in the upper mold. After the silicone cured (~40 min.), the finger was removed, and the excess silicone was severed from the finger body with a blade. The molding process can be seen in Fig. 3.

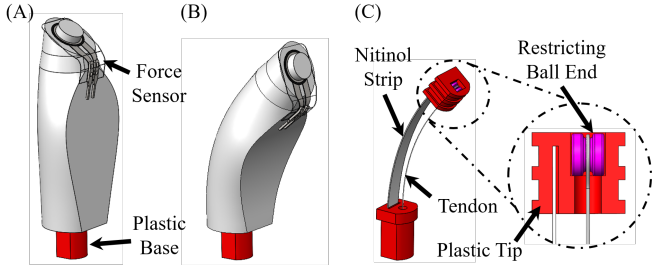

Fig. 2. (A) a side view of the finger in the unforced configuration, (B) deformed configuration caused by tendon retraction, and (C) the same forced configuration without the silicone body to illustrate the internal components. Note that the tendon was a guitar string. The ball end of the guitar string was restricted by interference in the plastic tip as shown in the detailed section view.

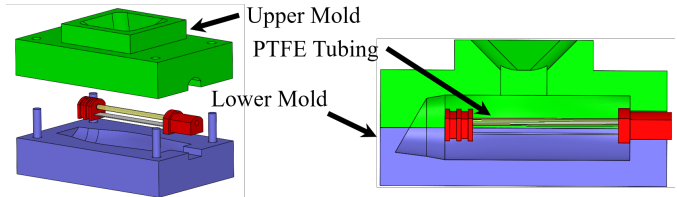

Fig. 3. Exploded view and assembled view of the molding process for the finger body with the finger backbone. Note that the nitinol strip provides the rigidity necessary to ensure that the plastic tip and plastic base are in line with one another.

## C. Force sensor integration

There is a wide variety of options that can be used to measure or estimate the task space force for soft robotics control [30, 45]. A common direct approach is using flexible resistive or capacitive sensors at the location that the force will be applied [46, 47]. In this study, force feedback was achieved by adhering a FlexiForce resistive force sensor (A301, Tekscan) to each fingertip using cyanoacrylate. This sensor was paired with a single power source non-inverting op-amp circuit using an MCP6004-I/P operational amplifier, as shown in Fig. 4 [48]. This circuit design was used due to its ability to (i) increase sensitivity with signal amplification, and (ii) provide a linear relationship between force and voltage, which is accomplished by the variable conductance of the FlexiForce sensor. A 100 kΩ potentiometer was used as a feedback resistor to further increase the sensitivity. To increase stability, a 47 pF capacitor was included in parallel with the feedback resistor. The supply voltage was 5V, and the reference voltage to the positive terminal of the amplifier was set to 3V using another 100 kΩ potentiometer.

The linear relationship between voltage and force was established using a simple calibration method based on a two-point linear relationship provided by: (i) a reference voltage of 3V with no applied force, and (ii) a voltage equal to 90% of the saturation voltage (5V) of the operational amplifier at 120% of the expected maximum applied force (3.9 N). The latter relationship was selected to provide the highest sensitivity without risk of saturation. Using this relationship, a desired force threshold for handling the berry can be related to a measured voltage.

According to the mechanical integration recommendation of Tekscan, a thin-cylindrical support (9.50 mm OD, 2.54 mm thick) was additively manufactured out of PLA and adhered to the sensing area of each FlexiForce sensor with double-sided tape. This support was used to ensure the force was entirely transmitted to the sensing area of the sensor as opposed to any other location on the finger body. The sensors were placed on the fingers such that the center of each sensor would achieve contact that was approximately normal to the berry's surface during finger actuation.

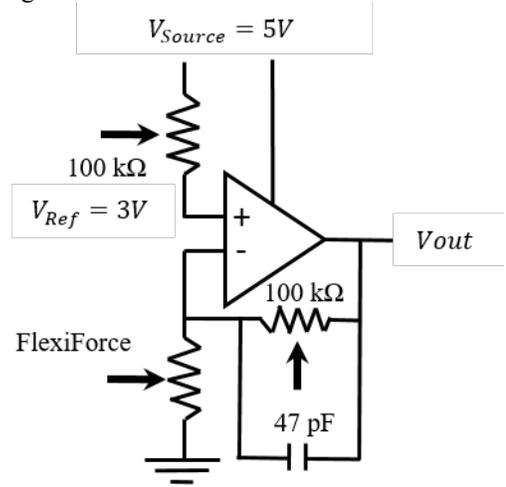

Fig. 4. FlexiForce circuit diagram. A 5V voltage source was supplied. A potentiometer was used to reduce this voltage to 3V, which is the reference voltage. A 47 pF capacitor was used with a 100 kΩ potentiometer to provide a highly sensitive, stable response. This potentiometer was trimmed to obtain the desired sensitivity.

## D. Finger Configuration

An important aspect of gripper design is the optimization of finger quantity, as well as their configuration on the palm. Optimization began with a qualitative investigation of the harvesting technique of skilled laborers through visual validation. It was found that laborers with smaller hands typically use four fingers to harvest blackberries, while harvesters with larger hands typically use only three fingers. Consequently, quantitative analysis was conducted to determine if the fourth finger of a laborer with smaller hands applied a negligible force. A custom-made force sensing system was developed to detect the forces applied by the fingertips of an experienced laborer using the FlexiForce A301 resistive force sensor for each finger, as shown in Fig. 5. These sensors used the aforementioned circuit shown in Fig. 4. Voltage data was measured using an Arduino Uno and sent through Bluetooth to MATLAB. In MATLAB, the voltage was converted to a force value based on the linear relationship determined in Section II C. A total of 2,160 blackberries were harvested using this system to determine the average force applied by each finger during the manual harvesting process. These results can be seen in Table 1.

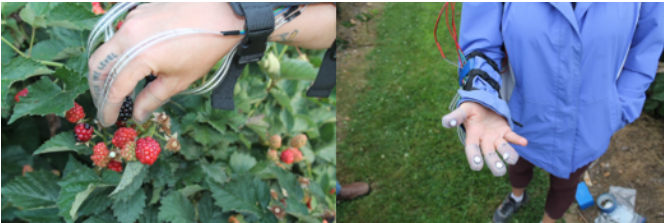

Fig. 5. Shown is the force-sensing apparatus for manual blackberry harvesting. Sensors were oriented to maximize contact with the berry surface during harvesting. Data recording and processing was conducted in a portable water-resistant case housed in a backpack.

TABLE 1: Average Forces per Finger for Manual Harvesting Blackberries

| Finger | Thumb | Index | Middle | Ring |
|---|---|---|---|---|
| Force [N] | 0.782 | 0.191 | 0.397 | 0.065 |

Table 1 indicates the need for three fingers and experimentally suggests that the fourth finger's applied load is negligible. It should be noted that the thumb had the highest average force of 0.782 N, followed by the middle finger with a value of 0.397 N. This can be explained by the anatomical opposition of the thumb to that of the middle finger when gripping small objects. Considering that a common mode of blackberry damage is drupelet deflation, it was decided that axis-symmetric spacing of the fingers is the optimum configuration to ensure an equal distribution of force is applied by each finger to the berry's surface.

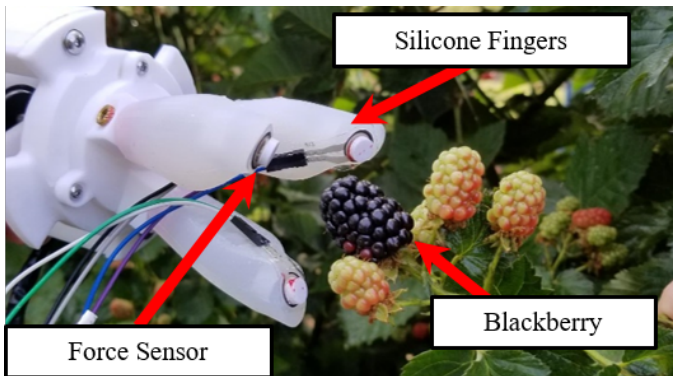

Fig. 6. The gripper is pictured in an unforced configuration prior to harvesting a ripe blackberry. Note the wide initial opening angle and the straightness of each finger.

### E. Soft Gripper Assembly

The fingers were 6.45 cm long and spaced 120° around the circumference of the palm. The fingers were offset by an angle of 20° from the vertical axis so that a maximum gripping diameter of 5.5 cm was provided, which accommodated the largest blackberry size and ensured that the outsides of the fingertips were within the mounting base profile.

Finger actuation was accomplished through tendon retraction via a stepper motor (17HS08-1004S, StepperOnline). The original stepper motor shaft was removed and replaced with a lead screw, which was secured with structural epoxy to the rotor. Rotational motion of the stepper motor was converted to linear translation through the press-fit of a lead screw nut into an additively manufactured linear bracket as shown in Fig. 7. The press-fit was sufficient in preventing dislocation of the lead nut from the linear bracket. The linear bracket extended radially to two linear rails and was affixed by means of a snap-fit to two linear bearings, which provided low-friction translation and constrained the linear bracket to strictly translational motion. The three tendons were each pre-tensioned and terminated below the linear bracket via knots located at the end of the tendons.

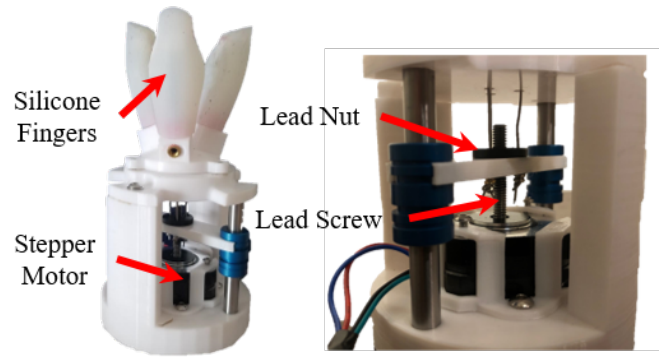

Fig. 7. Shown is the tendon-actuated gripper assembly. Due to the translucency of the silicone, observable is the gray nitinol strip and the red base of the finger backbone.

### F. Control System

The force feedback system was implemented in an Arduino Mega 2560 using a PD controller, which adjusted the stepper motor's rotational speed, thereby controlling tendon retraction speed. The desired speed was determined by first comparing the measured voltages from each of the fingers. The maximum measured voltage was then compared to a voltage corresponding to a maximum allowed force using the linear relationship between voltage and force. The resulting error and its differential determined the linear retraction speed. This operation was conducted using a global timer interrupt which directly reads from, and writes to, the Arduino Mega 2560's Timer1 register to run the stepping function exactly 10,000 times per second, while simultaneously taking input from the force sensors and outputting the desired speed using the PD controller. A proportional gain of 4000 and derivative gain of 125 were used. No integral gain was used to prevent potential overshoot. To manage the system's programmatic edge cases, the stepper motor turns itself off at the ends of its desired travel, as well as when the error is within the desired tolerance; as a result, the non-back-drivable lead screw prevents linear retraction from occurring through static friction. This feature acts to minimize the energy usage of the gripper, while also preventing damage to the gripper and to the berries, increasing the robustness of the system.

## III. Experimental Methods

### A. Force Control Reliability

Berry damage is primarily caused by excessive handling loads during the manual harvesting. We aim to evaluate the PD controller performance in order to maintain proper finger-berry contact force. The gripper was tested at several desired fingertip forces by comparing the desired force applied by the controller to the measured force of a high-resolution ATI force sensor (P/N: 9230-05-1311, ATI). The ATI sensor was constrained to a linear table and was positioned and oriented such that it was approximately parallel to the finger's force sensor in the finger's curled configuration, as shown in Fig. 8. The force control reliability test was conducted for eleven different desired force values ranging from 0.491 N to 1.472 N. This test was repeated eleven times for each respective force value, resulting in a total of 121 data points.

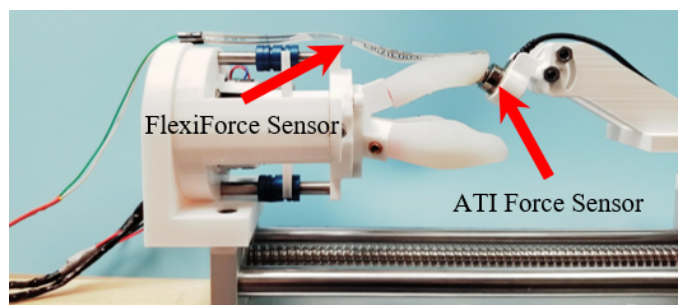

Fig. 8. Shown is the experimental setup to test the consistency of the PD controller. Note that the ATI force sensor was constrained such that it was approximately parallel to the FlexiForce sensor after finger retraction.

### B. Finger Actuation Characterization

The joint space (tendon retraction and force) was related to the task space (finger shape) by constructing an experimental setup that synchronously measured force and tendon retraction as the tendon was retracted. This was achieved using a linear sliding table that utilized an SFU1605 ball screw driven by a NEMA 23 stepper motor. The tendon wire was attached to a force sensor (Go Direct® Force and Acceleration Sensor, Vernier), which was constrained to the linear table. The finger assembly was fixed to the end of the linear rail using an additively manufactured bracket. A camera (5WH00002, Microsoft LifeCam Web Camera) was placed perpendicular to the longitudinal axis of the finger to record the resulting finger shape, as shown in Fig. 9. The camera was calibrated using the MATLAB Camera Calibration Toolbox [49]. The tendon was retracted from 0 to 10 mm in 0.50 mm increments. At each increment, the force measured from the Vernier force sensor was recorded in conjunction with an image of the finger being captured.

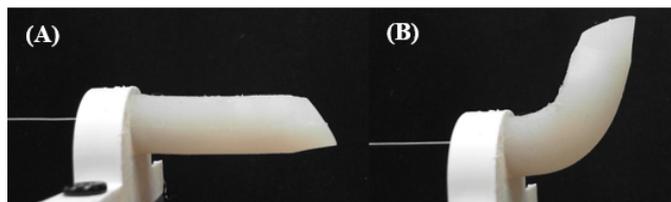

Fig. 9. The finger can be seen in a (A) 0mm configuration and (B) 10 mm configuration. The background was darkened to provide a sharp contrast for the imaging processing in post-analysis.

### C. Grasping Force (Max holding weight)

The grasping force test was used relate the maximum grasping force provided by the gripper in response to various geometric objects as they were pulled away at a constant speed. This was achieved using the linear sliding table in Sec. III-B. Six geometric solids were 3D printed and arranged in a seven configurations as shown in Fig. 10. Hooks were screwed into each of the geometric solids, permitting attachment to the force sensor (Go Direct® Force and Acceleration Sensor, Vernier). The force sensor was affixed to the linear table, and the gripper assembly was rigidly affixed to the end of the linear rail. The linear rail was oriented vertically and the force sensor was zeroed while holding each respective 3D-printed geometric shape to account for the weight of the object. The linear table was moved downward until the top of the shape was below the inner edge of the fingertips. The tendons of the fingers were then retracted until the fingers were coincident with the surface of the shape. The tendons of the gripper were then retracted by an additional 4 millimeters to apply a grasping force to the surface of each shape. The retention force was recorded while the linear table was linearly translated away from the gripper at a rate of 1 mm/s until the object was no longer constrained by the gripper. A similar procedure was repeated for only the 3D printed sphere and cylinder at tendon retractions ranging from 1 to 7mm.

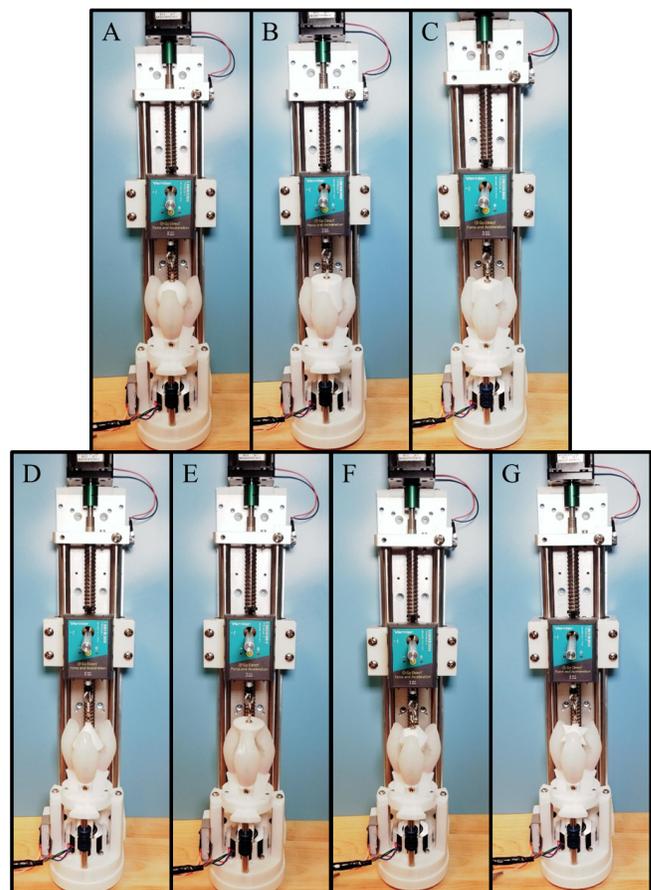

Fig. 10. Shown are the various geometric shapes used to test the gripping retention force, including a sphere, cylinder, cube, upright cone, inverted cone, icosahedron, and stellated dodecahedron, labeled respectively from A to G. All objects had major dimensions (diameter, height, length) of 3.81 cm, with the exception of the cube, which had a side length of 2.54 cm.

### D. Fingertip Force Characterization

The normal force applied by the gripper to objects of various diameter was characterized by constructing an experimental setup that measured the force needed to push and pull a 3D printed cone with a 15° draft angle between the gripper's fingers at a constant velocity. A cone was chosen for this experiment because it enabled a continuous and axis-symmetric method of measuring a range of object diameters, where the object diameter grasped by the gripper is a function of linear translation. The 3D printed cone was split into two truncated pieces to allow a greater range of gripping diameters. The experimental setup consisted of using the aforementioned linear rail assembly. A force sensor (Go Direct® Force and Acceleration Sensor, Vernier) was affixed to the linear table via a metal rod, and the gripper assembly was rigidly affixed to the end of the linear rail. For measuring the pulling force, a procedure similar to Section III-C was used with truncated cones as the objects. The diameter at the base of the fingertip was marked as the initial diameter and the object was pulled from the gripper at a speed of 1mm/s until the gripper was no longer in contact with the object. The force sensor output was recorded during the entirety of the object's translation.

For measuring the pushing force, the hooks were removed from the truncated cones, and each part was placed in the closed gripper at the tendon retraction to be tested. The diameter at the base of the fingertip was marked as the initial diameter. The truncated cones were then pushed into the fingers of the gripper via the linear rail at a speed of 1mm/s, and the insertion force was measured with the force sensor.

Similar pushing and pulling procedures were repeated for both truncated cones at tendon retractions ranging from 5 to 9 mm in 0.5mm increments. Because the measured force and the position of the cone are related to time, the cross-sectional diameter of the cone can be related to the measured force based on the draft angle of the cone and its position with respect to the gripper.

### E. Grasping Versatility

Although the gripper was designed specifically for berry harvesting, it has potential to be used to handle objects of activities of daily living (ADL). As a result, gripper versatility was tested on a wide variety of object sizes, geometries, and material properties. This included a variety of fruits and vegetables, glass and plastic bottles, and other rigid and non-rigid objects. This was done by simply placing the object to be tested within the gripper's workspace and retracting the tendons of the gripper until an adequate amount of force was applied to prevent the object from falling.

### F. Field Test (harvest reliability, speed, and damage)

The final test of this gripper's efficacy was an intensive field test that involved harvesting a total of 240 blackberries at a private pick-your-own berry farm in Fayetteville, AR. The gripper was manually positioned and oriented by a graduate student and the berry was harvested once it was within the workspace of the gripper. Three different desired fingertip force thresholds were used based on our preliminary manual harvesting data. Sixty berries were harvested at a desired fingertip force value of 0.59 N, 0.69 N, and 0.78 N, respectively. Additionally, sixty blackberries were also harvested with the force sensors removed, using a tendon retraction of 4 mm; although this provided no force feedback, it removed the rigid cylindrical support from the fingertip, resulting in exclusively soft, compliant contact with the berries. The blackberries for this evaluation had a mass of roughly 8 g each with a length of 30 mm and a width of 21 mm. For each tip force level, the blackberries were harvested and placed in plastic, vented clamshell containers (20 berries per clamshell), then stored at 2°C for 21 days to evaluate RDR.

## IV. RESULTS AND DISCUSSION

### A. Force Control Reliability

Using the calibration method described in Section II-C, the linear relationship provided by the circuit indicated a slope of $5.232 V/N$. Using this relationship, a desired threshold voltage was set for the controller based on the desired force values. For each desired force value, the force from the ATI force sensor was recorded for the 11 trials. This force was converted to voltage using the calibrated linear regression. A plot of the calibrated linear regression (hyphenated blue line) and the resulting force values (black error bars) of the experimental results can be seen in Fig. 12. Due to the novelty of robotic blackberry harvesting, there remain many unknown design constraints, such as the required force accuracy to prevent berry damage; however, the average error between the desired force and the applied force to the ATI force sensor was 0.046 N, indicating high controller reliability.

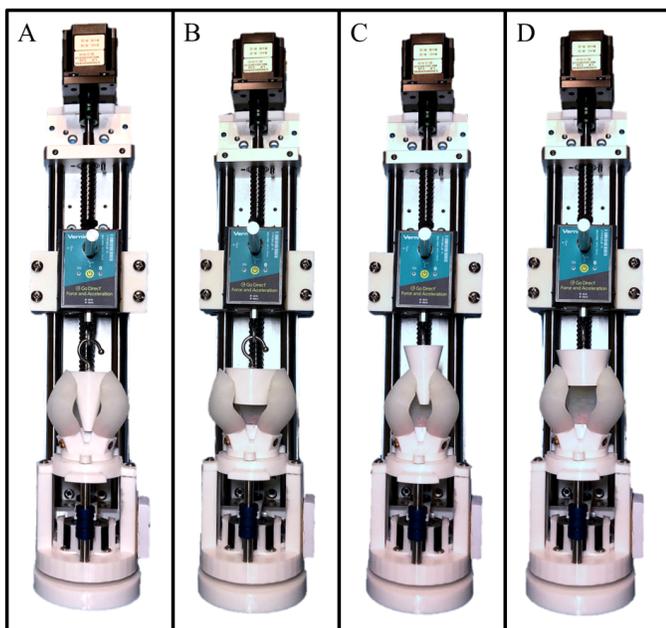

Fig. 11. Shown is the experimental setup for the pulling (A, B) and pushing (C, D) portions of the fingertip force characterization test.

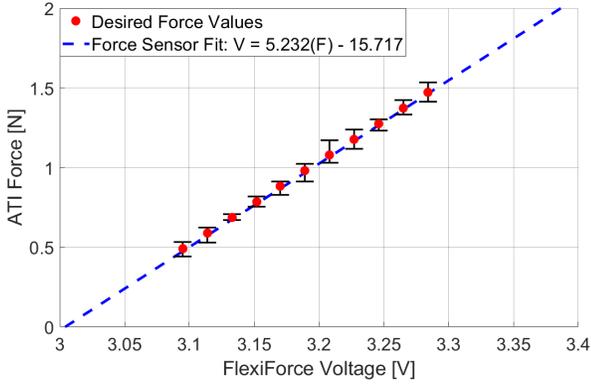

Fig. 12. Depicted graphically are the desired force values (red dots) overlaid on the linear calibration line from Section II C (hyphenated blue lines). The error bars represent the experimentally measured results.

### B. Finger Actuation Characterization

Using the procedure discussed in Section III-B, at each tendon retraction, image processing was used to fit a circle to the inner surface of the finger, and the curvature associated with each configuration of the finger was calculated and recorded. Fig. 13 depicts the curvature with respect to each force (A) and tendon retraction (B). A quadratic curve was fit to each data set to provide a relationship between the independent variables to the curvature for the procedures in the following experiments. Fig. 13A depicts every recorded data point, with an average error between the experimental curvature and the curve-fitted curvature of 5.03 %. Fig. 13B depicts the relationship between tendon retraction and the resulting average curvature, with an average percent error between the experimental curvature and the curve fitted curvature of 5.50%.

**(A)**

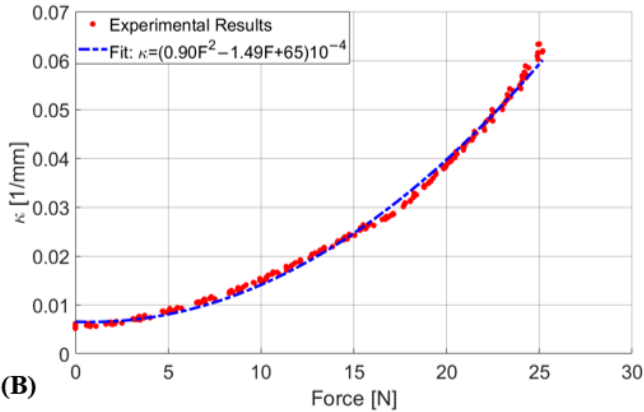

**(B)**

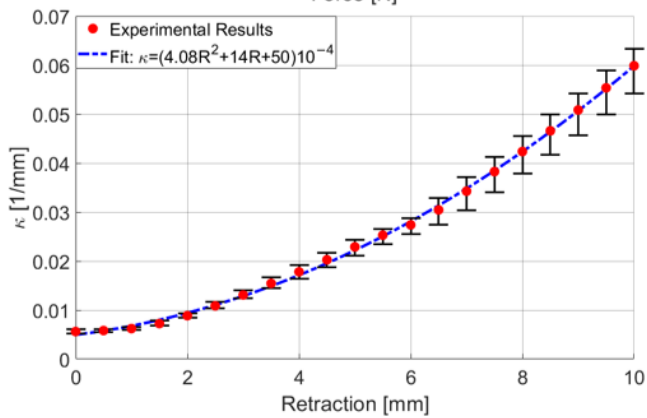

Fig. 13. (A) shows the relationship between tendon force and the resulting curvature of the inner face of the finger. (B) shows the relationship between the tendon retraction and the resulting curvature.

### C. Gripping Force (Max holding weight)

The overall maximum force sustained by the gripper was 18.94 N, which was accomplished with the stellated dodecahedron, as shown in Fig. 14 This shows that the gripper is capable of handling a payload of nearly 2 kg, noting that objects possessing sharp edges (stellated dodecahedron, cone) were retained at higher loads than objects with smooth features (cylinder, sphere). This is clearly captured by the measured retention force of the sphere and cylinder at a tendon retraction of 4 mm as shown in Fig. 14. It was found that at 4 mm, the gripper can handle a 0.28 kg payload for a 3.81 cm diameter spherical object and a 0.38 kg payload for a 3.81 cm diameter cylindrical object, both of which are less than 20% of the payload of the stellated dodecahedron at 4 mm tendon retraction (1.93 kg). It is speculated that these sharp edges provide the gripper with a form of mechanical advantage by allowing the tendon-actuated silicone body to conform to the sharp edges.

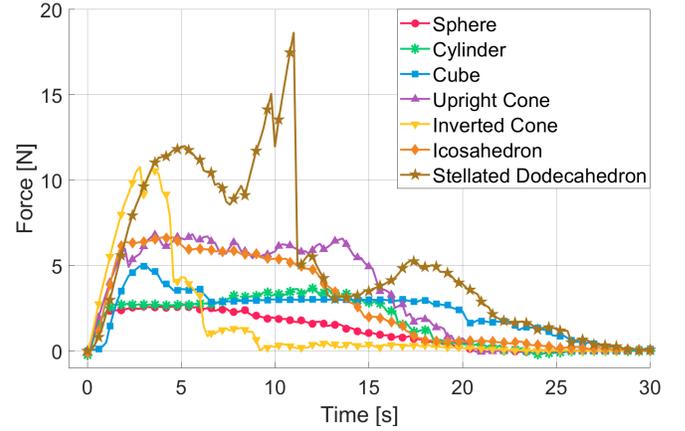

Fig. 14. Depicted are the results from the gripper retention force test for each of the seven loading scenarios at a tendon retraction of 4mm. The maximum retention force provided by the gripper was 2.75 N for the sphere, 3.72 N for the cylinder, 4.96 N for the cube, 6.84 N for the upright cone, 10.84 N for the inverted cone, 6.65 N for the icosahedron, and 18.94 N for the stellated dodecahedron.

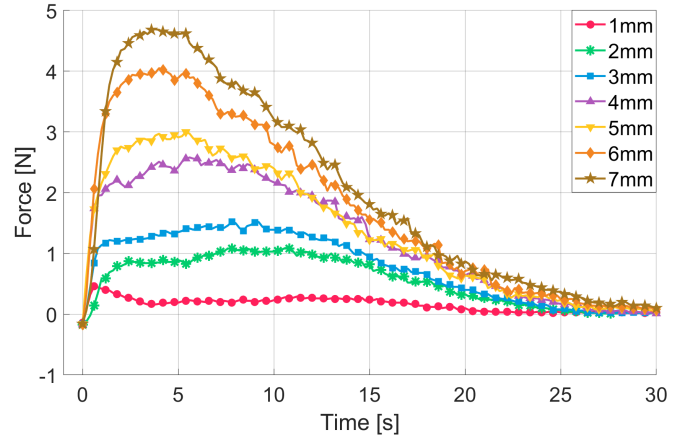

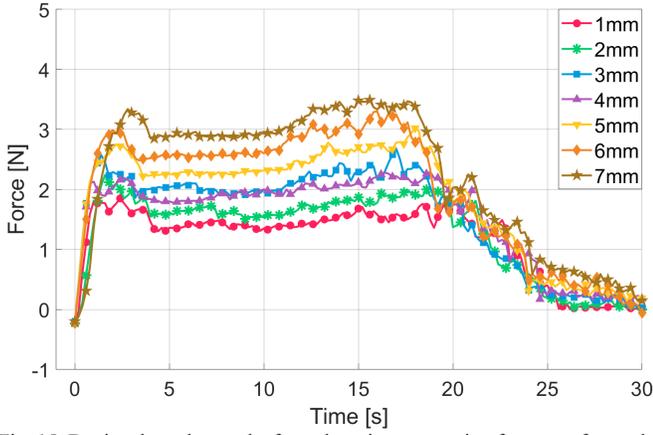

Fig. 15. Depicted are the results from the gripper retention force test for tendon retraction lengths of 1-7mm when gripping the sphere (top image) and cylinder (bottom image). The maximum retention force provided by the gripper was 4.71 N for the sphere and 3.50 N for the cylinder.

### D. Fingertip Force Characterization

The relationship between tendon retraction, object diameter, and fingertip force is calculated by considering the free body diagrams in the pushing and pulling scenarios, as shown in Fig. 16. By summing the forces in the direction the cone is traversing, we obtain equation (1) and (2) for the pulling and pushing configuration, respectively.

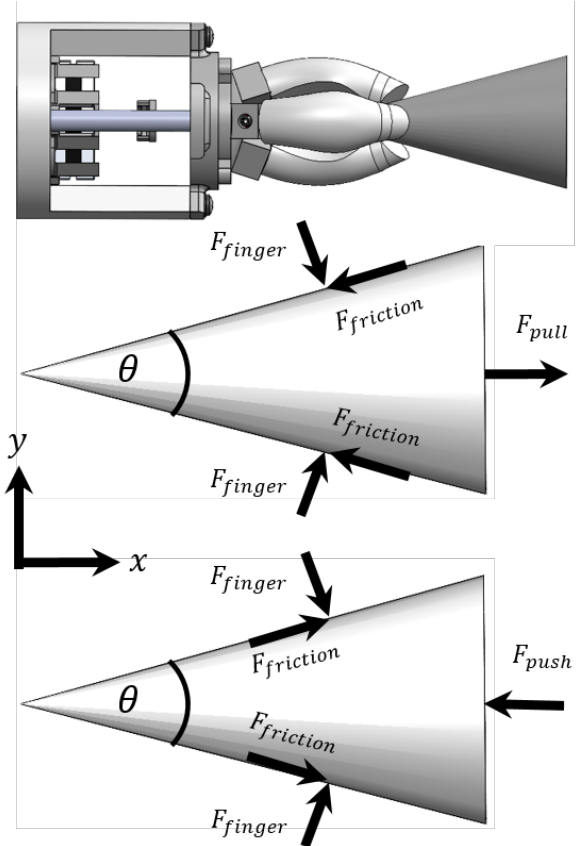

Fig. 16. Depicted is a 2D representation of the experimental setup, which is equivalent to the 3D setup due to the axial symmetry of the cone and gripper.

$$\sum F_{x,pull}: \ F_{pull} = n(-F_{finger}\sin\left(\tfrac{\theta}{2}\right) + \mu F_{finger}\cos\left(\tfrac{\theta}{2}\right)) \quad (1)$$

$$\sum F_{x,push}: \ -F_{push} = n(-F_{finger}\sin\left(\tfrac{\theta}{2}\right) - \mu F_{finger}\cos\left(\tfrac{\theta}{2}\right)) \quad (2)$$

where $F_{finger}$ is the force applied by each fingertip normal to the cone, $n$ is the number of fingers on the gripper, $\theta$ is the cone angle, $\mu$ is the coefficient of kinetic friction, and $F_{push}$ and $F_{pull}$ are the forces measured in the pushing and pulling tests, respectively. By adding (1) and (2), and solving for $F_{finger}$, we obtain:

$$F_{finger} = \frac{F_{push} - F_{pull}}{2n*\sin\left(\tfrac{\theta}{2}\right)} \quad (3)$$

Knowing the result of Section III-D, the normal force applied by the gripper at a given tendon retraction can be solved for numerous object diameters to obtain statistically significant results.

Experimental results from the two truncated cones were normalized to produce continuous data at gripping diameters ranging from 9 to 47 mm. The results can be seen in Fig. 17. The maximum fingertip force, which was calculated from the pushing and pulling forces, was determined to be 4.92 N at an object diameter of 47 mm and a tendon retraction of 9 mm.

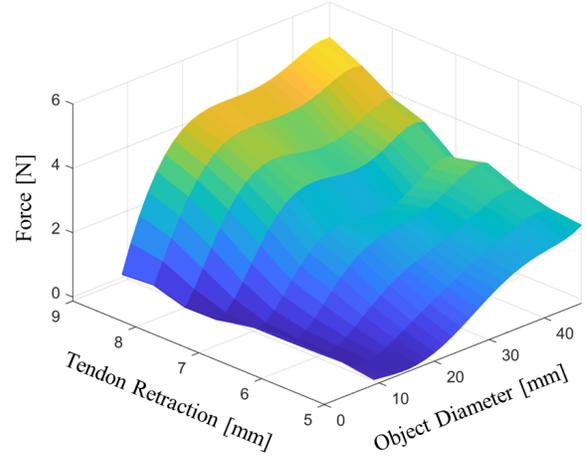

Fig. 17. Depicted are the calculated gripping forces for tendon retraction lengths of 5-9mm.

### E. Grasping Versatility

The grasping versatility of the gripper is illustrated by the handling of objects that are a part of activities of daily living as seen in Fig. 18. These objects were selected such that there was a variety in geometry, stiffness, weight, and compliance for testing the efficacy of the gripper. The ability of this gripper to delicately grip soft objects is clearly represented by Fig. 18 I, where the gripper is used to hold a pastry. Conversely, the strength of the gripper is displayed by its ability to hold a can of beans and a jar of pears as shown in Fig. 18 B and Fig. 18 E, respectively. Finally, the ability of the gripper to hold compliant objects is verified by the handling of the T-shirt in Fig. 18 H and the bag of chips of Fig. 18 L.

It should be noted that the internal plastic tip used to terminate the tendon within the silicone fingers provided useful mechanical advantage as illustrated in Fig. 18 B and E. For these objects, the gripper fingers are forced into an angle greater than the initial offset angle of 20°, which could potentially cause an inability to grasp these large diameter objects without assistance; however, as the tendon is retracted, the internal plastic component essentially wraps around the lip of the can and the lid of the jar, creating inherent mechanical advantage. This is similar in the way that human fingers can curl around a small object, reducing the need to rely on frictional grasping.

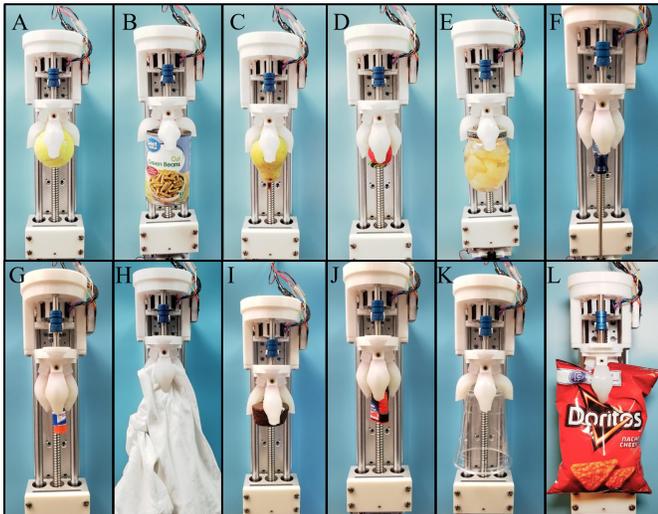

Fig. 18. Shown is the gripper being used to handle various objects. These objects are a tennis ball (A), a can of beans (B), a pear (C), a strawberry (D), a jar of pears (E), a screwdriver (F), a glue stick (G), a t-shirt (H), a pastry (I), a container of super glue (J), an upside-down plastic cup (K), and a bag of chips (L).

### F. Field Test

The results of the field test harvesting can be seen in Table 2. RDR [%] was the percentage of 60 berries that were harvested and stored in a clamshell at 2° C for 21 days that succumbed to RDR a post-harvest disorder where the drupelets on the blackberry turn from black to red. Reliability [%] refers to the number of berries harvested divided by the total number of harvesting attempts, in this case 60 per trial. The harvest time refers to the amount of time required to approach a berry, grasp it, remove it from the plant, and place it in a separate holding vessel.

In the force feedback tests, reliability, speed, and RDR [%] were inversely related to the applied force. It should be noted that harvest time can be reduced by tuning the PD gains to decrease response time. The blackberries that were harvested manually and with FB 1 did not have any RDR. Interestingly, when using no force feedback with the gripper, RDR [%] and response time are reduced when compared to the force feedback method with a desired force of 0.78 N. However, reliability suffers as a consequence of the lack of active force feedback. These results indicate that the cylindrical support attached to the FlexiForce sensor causes damage to the berry surface. To address this issue, the force sensing interface should be removed from the fingertips, and a mapping between the joint space force (gripper force), task space force (tendon force), and berry diameter should be performed.

TABLE 2: Results of Harvesting Blackberries with Varying Force Feedback

| Parameter | FB 1[i] | FB 2 | FB 3 | No FB[ii] | Hand[iii] |
|---|---|---|---|---|---|
| Desired Force [N] | 0.59 | 0.69 | 0.78 | N/A | N/A |
| Reliability [%] | 77.92 | 86.96 | 95.24 | 85.71 | 100.00 |
| Harvest Time [s] | 8.10 | 7.30 | 4.80 | 3.50 | 1.40 |
| Red Drupelet Reversion [%] | 0.00 | 8.00 | 16.00 | 0.00 | 0.00 |

i. FB refers to a gripper test that utilized the force sensors to provide force feedback during harvesting.
ii. No FB refers to a gripper test that had the force sensors removed during harvesting. A maximum tendon retraction of 4 mm was used for each berry.
iii. Hand refers to a test where berries were harvested by hand.

### V. Conclusions

This paper presents a novel tendon-driven gripper for the harvesting of delicate, plant-ripened berries, as well as objects used in ADL. This gripper was capable of handling a maximum payload of nearly 2 kg (19 N). Force feedback was provided through the implementation of a flexible resistive force sensor and an operational amplifier. As a result, each finger was able to apply a desired force as low as 0.49 N to as high as 1.47 N with an average error of 0.046 N. Field harvesting tests indicate that this gripper is capable of equal performance when compared to manual harvesting, especially at the lower force feedback with respect to RDR, although RDR remained very low for all force feedback levels. All force feedback levels maintained a harvesting speed of under 9 seconds per berry.

Future work includes mapping joint space force to task space force, i.e. measuring fingertip force indirectly by measuring the tension in the tendons to reduce berry damage caused by the plastic cylindrical force sensor supports attached to the gripper's fingertips. Additionally, the gripper's response to external forces, such as canes and leaves, will be investigated in order to develop a more robust system. The end goal will be to attach the gripper to a custom-made mobile robot chassis that utilizes simultaneous localization and mapping to autonomously harvest blackberries using the integration of novel image processing into the system.

### Acknowledgment

This work was supported by University of Arkansas Chancellor's Fund for Innovation and Collaboration.

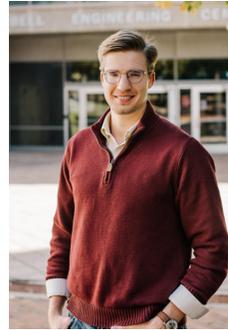

**Anthony L. Gunderman** was born in Batesville, AR, USA in 1997. He received a B.S. in mechanical engineering from the University of Arkansas, Fayetteville, AR, USA in 2019. He is currently pursuing a master's degree in electrical engineering and a Ph.D. in mechanical engineering at the University of Arkansas, Fayetteville, AR, USA in 2019.

In 2017 and 2018, he participated in two cooperative education stints with L3-Technologies as a Design Engineer. Since 2019, he has been conducting research in medical and soft robotics with Dr. Yue Chen in the Mechanical Engineering Department at the University of Arkansas, Fayetteville, AR, USA.

Mr. Gunderman is a member of the mechanical engineering honor society Pi Tau Sigma. Additionally, he is a national member of ASME.

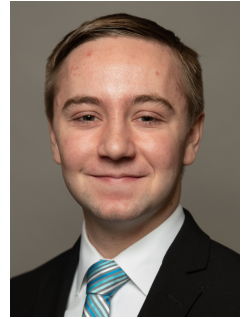

**Jeremy A. Collins** was born in Pelsor, AR, USA in 1999. He is currently pursuing a B.S. in mechanical engineering from the University of Arkansas, Fayetteville, AR, USA.

In 2018, he was employed as an Application Development Intern with J.B. Hunt Transport Services, and in 2019, he was employed as a Mechanical Engineering Intern with Marshalltown Company.

Since 2019, he has been conducting research in medical and soft robotics with Dr. Yue Chen in the Mechanical Engineering Department at the University of Arkansas, Fayetteville, AR, USA.

Mr. Collins is a national member of ASME and a member of the IEEE Power Electronics Society.

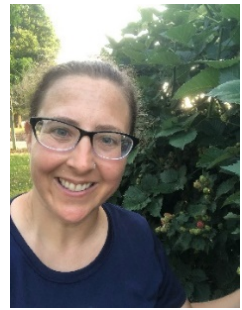

**Andrea Myers** received a B.S. in food and culinary science from University of Arkansas, Fayetteville, AR, USA in 2019. She is currently pursuing a master's degree in food science at the University of Arkansas, Fayetteville, AR, USA. Her current research is on flavor and handling of fresh-market blackberries.

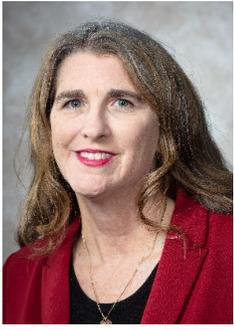

**Renee Threlfall** received a B.S. in microbiology from the University of Arkansas, Fayetteville, AR, USA in 1990, followed by a M.S. and Ph.D. in 1992 and 1997, respectively. She is currently a research scientist in the Food Science Department at the University of Arkansas System Division of Agriculture. Her research efforts at the University of Arkansas are focused on specialty crops with expertise in processing and postharvest of fruits (grapes, blackberries, peaches, strawberries, etc.). Dr. Threlfall is a member of the American Society of Enology and Viticulture (ASEV), ASEV-Eastern Section, the American Wine Society, and the American Society for Horticultural Science, and American Society of Brewing Chemists. She serves as the Administrator for the ASEV-Eastern Section. Dr. Threlfall is on the Extension and Outreach Committee for the National Grape Research Alliance.

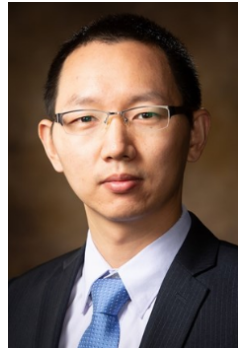

**Yue Chen** (M' 19) received the B.S. in vehicle engineering from Hunan University, Hunan, China, in 2010, M.Phil. in mechanical engineering from Hong Kong Polytechnic University, Hung Hom, Hong Kong, in 2013, and Ph.D. in Mechanical Engineering, Vanderbilt University, Nashville, TN, USA, in 2018. He started the assistant professor position in 2018 at the Department of Mechanical Engineering, University of Arkansas, Fayetteville, AR, USA. His current research interests include medical robotics and soft robots.